\documentclass{article}
\usepackage[main, final]{neurips_2026}
\usepackage{graphicx}
\usepackage[utf8]{inputenc} 
\usepackage[T1]{fontenc}    
\usepackage{hyperref}       
\usepackage{url}            
\usepackage{booktabs}       
\usepackage{amsfonts}       
\usepackage{nicefrac}       
\usepackage{microtype}      
\usepackage[dvipsnames,table,xcdraw]{xcolor}
\usepackage{xspace}
\usepackage{pifont}
\usepackage{multirow}
\usepackage{amsmath}
\usepackage{fvextra}
\usepackage{etoolbox}
\usepackage[capitalize,noabbrev]{cleveref}

\definecolor{color5}{HTML}{006795}
\hypersetup{
colorlinks   = true, 
urlcolor     = blue, 
linkcolor    = color5, 
citecolor   = color5 
}

\newcommand{\benchmark}{\textsc{MINTEval}\xspace}
\newcommand{\qwen}{Qwen3.6-35B-A3B\xspace}
\newcommand{\gemini}{Gemini-3.1-Flash-Lite\xspace}
\newcommand{\horizonbench}{HorizonBench\xspace}
\newcommand{\babi}{bAbI\xspace}
\newcommand{\simple}{\texttt{Simple}\xspace}
\newcommand{\history}{\texttt{History}\xspace}
\newcommand{\order}{\texttt{Ordering}\xspace}
\newcommand{\counting}{\texttt{Counting}\xspace}
\newcommand{\multihop}{\texttt{Multihop}\xspace}

\newcommand{\greencmark}{\textcolor{ForestGreen}{\ding{51}}}
\newcommand{\redxmark}{\textcolor{red}{\ding{55}}}

\definecolor{FullBaseline}{HTML}{F3850F}
\definecolor{RAGBaseline}{HTML}{519872}
\definecolor{MemBaseline}{HTML}{2782ed}
\definecolor{OurColor}{HTML}{36aa70}

\usepackage[most]{tcolorbox}     
\definecolor{ExampleBg}{HTML}{ffffff}
\definecolor{ExampleTitle}{HTML}{629677}
\newcounter{example}

\tcbset{examplestyle/.style={
    enhanced jigsaw,
    breakable,                  
    colback=ExampleBg,          
    colframe=ExampleTitle,      
    colbacktitle=ExampleTitle,  
    coltitle=white,             
    fonttitle=\bfseries,        
    arc=5pt,                    
    boxrule=0.5pt,              
    before skip=10pt,           
    left=6pt, right=6pt,
    top=3pt, bottom=3pt,
    toptitle=3pt, bottomtitle=0pt,
}}

\newenvironment{example*}[1][]{%
    \ifstrempty{#1}%
        {\begin{tcolorbox}[examplestyle]}%
        {\begin{tcolorbox}[examplestyle, title={#1}]}%
}{%
    \end{tcolorbox}%
}

\DefineVerbatimEnvironment{PromptText}{Verbatim}{%
  fontsize=\footnotesize,%
  breaklines=true,%
  breaksymbolleft={},%
  breaksymbolsepleft=0pt,%
  breakindent=0pt,%
  breakautoindent=false%
}

\definecolor{cellRangeA}{RGB}{255, 215, 215}   
\definecolor{cellRangeB}{RGB}{255, 240, 240}   
\definecolor{cellRangeC}{RGB}{208, 255, 203}   
\definecolor{cellRangeD}{RGB}{174, 252, 172}   
\definecolor{cellRangeE}{RGB}{166, 223, 166}  

\newcommand{\cv}[1]{%
  \ifdim #1pt<20pt \cellcolor{cellRangeA}#1%
  \else\ifdim #1pt<40pt \cellcolor{cellRangeB}#1%
  \else\ifdim #1pt<60pt \cellcolor{cellRangeC}#1%
  \else\ifdim #1pt<80pt \cellcolor{cellRangeD}#1%
  \else \cellcolor{cellRangeE}#1%
  \fi\fi\fi\fi}

\usepackage{titlesec}

\titlespacing*{\paragraph}
{0pt}   
{0pt}   
{1em}   

\title{\benchmark: Evaluating Memory under Multi-Target Interference in Long-Horizon Agent Systems}

\author{
\textbf{Hyunji Lee}\textsuperscript{1}\thanks{Equal contribution; order decided by a coin flip. Corresponding authors: \texttt{\{hyunjil,cychen\}@cs.unc.edu}} \quad
\textbf{Justin Chih-Yao Chen}\textsuperscript{1\ensuremath{\ast}} \quad
\textbf{Joykirat Singh}\textsuperscript{1} \quad
\textbf{Zaid Khan}\textsuperscript{1} \quad \\ 
\textbf{Elias Stengel-Eskin}\textsuperscript{2} \quad
\textbf{Mohit Bansal}\textsuperscript{1} \\
\textsuperscript{1}UNC Chapel Hill \quad
\textsuperscript{2}The University of Texas at Austin 
}

\begin{document}

\maketitle

\begin{abstract}
Agents in real-world settings operate over long and evolving horizons, where information is repeatedly updated and may interfere across memories, requiring accurate recall and aggregated reasoning over multiple pieces of information.
However, existing benchmarks focus on static, independent recall and fail to capture these dynamic interactions between evolving memories. 
In this paper, we study how current memory-augmented agents perform in realistic, interference-heavy, long-horizon settings across diverse domains and question types.
To this end, we introduce \benchmark (Long-Horizon \underline{\vphantom{g}M}emory under \underline{\vphantom{g}INT}erference \underline{\vphantom{g}Eval}uation), an analytical benchmark which features 
(1) long, highly interconnected contexts with frequently updated information that induces substantial interference,
(2) diverse domains (state tracking, multi-turn dialogue, Wikipedia revisions, and GitHub commits), enabling evaluation of domain generalization, 
and (3) diverse question types that assess robustness to interference, including \textit{(i) single-target recall} tasks requiring retrieval of a specific target from long contexts, and \textit{(ii) multi-target aggregation} tasks requiring reasoning over multiple relevant pieces of information.
Overall, \benchmark contains 15.6k question-answering pairs over long-horizon contexts averaging 138.8k tokens and extending up to 1.8M tokens per instance.
We evaluate seven representative systems, including vanilla long-context LLMs, retrieval-augmented generation methods, and memory-augmented agent frameworks.
Across all systems, we observe consistently low performance (avg. 27.9\% accuracy), especially on questions requiring aggregated reasoning over multiple pieces of evidence. 
Fine-grained analysis shows that performance is primarily limited by retrieval and memory construction capabilities.
Furthermore, current memory systems struggle to recall and reason over earlier facts that are later revised or interfered with by subsequent context, with performance degrading as the number of intervening updates increases. 
These findings highlight the need for more robust memory management systems for dynamic, long-horizon environments across varying domains\footnote{Code and data are available at \color{magenta}{\url{https://github.com/amy-hyunji/MINTEval}}}.
\end{abstract}

\section{Introduction}

\begin{figure*}
    \centering
    \includegraphics[width=\linewidth]{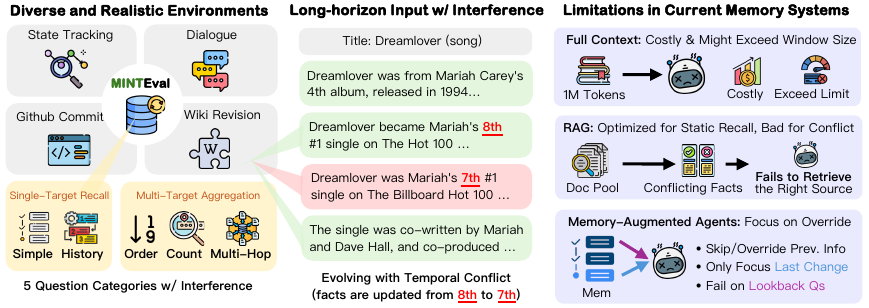}
    \caption{
    \textbf{Left:} \benchmark{} spans four realistic domains: state tracking, dialogue, GitHub commits, and Wikipedia revisions, with five question categories probing different aspects of memory behavior. \textbf{Middle:} The contexts are inherently dynamic and continuously evolving, naturally creating frequent destructive interference. \textbf{Right:} Existing memory systems show distinct failure modes: (1) full-context methods are computationally expensive and exceed context limits, (2) RAG systems often retrieve incorrect evidence due to conflicting information, and (3) memory-augmented agents overemphasize recent information and underuse historical context, hurting lookback-style queries.}
    \label{fig:fig1}
\end{figure*}

Memory-augmented agents powered by large language models (LLMs) are increasingly being developed to support a variety of tasks~(e.g., long-horizon tasks~\citep{huang2026rethinking,gutierrez2025ragmemorynonparametriccontinual,hu-etal-2025-hiagent} and lifelong learning~\citep{11328884,zheng2025lifelongagentbench,liu2025memverse}), where information continuously accumulates over time~\citep{ong2025towards, kim2026largelanguagemodelsup}.
In many real-world settings, newly acquired information does not fully overwrite prior information, but instead revises or builds upon existing states. 
For example, software systems and documents evolve through successive revisions that introduce new features or modify existing syntax and behaviors. In such settings, users may query specifications from older versions or compare differences across revisions when migrating to newer releases.
Similarly, during long-term interactions with conversational agents, users continuously provide new information across multiple interactions that may reinforce, modify, or contradict earlier preferences or personal attributes~\citep{Chen_2026, mehri2026multisessioncollablearninguserpreferences}.
Users may ask about facts or preferences they no longer recall, or expect agents to respond consistently with preferences expressed throughout past interactions.
These real-world settings require agents not only to preserve information over time, but also to understand how newly acquired information relates to prior states, enabling agents to recall and aggregate information across interactions rather than simply overwrite existing memories.
However, as information accumulates over long horizons, \textit{interference}\footnote{Here, \textit{interference} encompasses both proactive interference, where old memories affect encoding of new information, and retroactive interference, where new information overwrites existing ones.} naturally emerges, which is a well-studied phenomenon in human memory~\citep{underwood1957interference, ANDERSON1996237} (Fig.~\ref{fig:fig1}, middle) where previously stored and newly acquired information interact and conflict with one another, making retrieval and reasoning over past information challenging.

A simple solution to answering such questions with long horizon context is to include all available context in the input, especially as model context lengths have grown substantially in recent years~\citep{team2024gemini,qwen3technicalreport}, but this remains inefficient and often exceeds practical context length limits~\citep{kim2026largelanguagemodelsup, wang2025memalpha}. 
To address this, memory-augmented agents have been proposed~\citep{xu2025amemagenticmemoryllm,huo2026atommemlearnabledynamic,packer2024memgptllmsoperatingsystems,zhou2025mem1learningsynergizememory}, which store, update, and retrieve information over time while preserving consistency. These approaches have demonstrated stronger and more robust performance than both naive full-context usage and standard retrieval-augmented generation~(RAG).
However, important gaps remain in understanding how memory-augmented agents perform in real-world settings, as shown in Fig. \ref{fig:fig1} (right). 
As shown in Table~\ref{table: comparison} (\textsc{Interdep.} and \textsc{Interference} columns), existing memory benchmarks often focus on long-horizon inputs composed of largely independent events with sparse interactions~(e.g., concatenating unrelated contexts into a single long sequence~\citep{hu2026evaluatingmemoryllmagents, wang2025memalpha}), failing to capture the dense and evolving \emph{interference-heavy} contexts observed in real-world memory. 
Also, existing benchmarks~\citep{wang2025memalpha, wan2025storybenchdynamicbenchmarkevaluating} primarily focus on recall of recent information, while overlooking \emph{long-range lookback}\footnote{By \emph{long-range lookback}, we mean queries about information from much earlier in the interaction history rather than the latest state, e.g., if a person moved ten times, it may ask where they lived after the third move instead of where they live now.}~(\textsc{LookBack}) and reasoning tasks that require aggregating multiple relevant targets~(\textsc{Aggr.}). 
Moreover, existing benchmarks are often focused on specific domains, particularly conversational environments~\citep{tavakoli2026milliontokensbenchmarkingenhancing, wu2025longmemevalbenchmarkingchatassistants}, thereby failing to evaluate \emph{domain generalization}~(\textsc{M-Domain}).

\begin{table*}[t!]
\centering
\small
\caption{
Comparison of \benchmark with prior memory benchmarks. We categorize benchmarks by (1) input context properties: interdependent inputs (\textbf{Interdep.}), dense interference ($\geq$10 depth; \textbf{Interference}), and multi-domain coverage (\textbf{M-Domain}); and (2) question properties: multi-target aggregation (\textbf{Aggr.}) and lookback to earlier context (\textbf{LookBack}). Further benchmark comparisons in Appendix~\ref{app:more_benchmarks}.
}
\begin{tabular}{lccc|cc}
\toprule
& \multicolumn{3}{c}{Input Context} & \multicolumn{2}{c}{Question Type} \\
 \cmidrule{2-4} \cmidrule{5-6}
Benchmark & Interdep. & Interference & M-Domain & Aggr.& LookBack \\
\midrule 
MemoryAgentBench~\citep{hu2026evaluatingmemoryllmagents} & \redxmark & \redxmark & \greencmark & \greencmark & \redxmark\\
Mem-$\alpha$~\citep{wang2025memalpha} & \redxmark & \redxmark & \greencmark & \redxmark & \redxmark\\
Locomo~\citep{maharana2024evaluating} & \greencmark & \redxmark &\redxmark & \greencmark & \redxmark\\
LongMemEval~\citep{wu2025longmemevalbenchmarkingchatassistants} & \greencmark & \redxmark & \redxmark & \greencmark & \redxmark \\
BEAM~\citep{tavakoli2026milliontokensbenchmarkingenhancing} & \greencmark & \redxmark& \redxmark & \greencmark & \redxmark \\
StoryBench~\citep{wan2025storybenchdynamicbenchmarkevaluating} & \greencmark & \redxmark & \redxmark & \redxmark & \redxmark \\
OAKS~\citep{kim2026largelanguagemodelsup}  & \greencmark & \redxmark & \redxmark  & \greencmark & \redxmark \\
\rowcolor{blue!7}\benchmark (Ours) & \greencmark & \greencmark & \greencmark & \greencmark & \greencmark \\
\bottomrule
\end{tabular}
\label{table: comparison}
\end{table*}

To evaluate how memory-augmented agents perform under such settings, we introduce an analytical benchmark, \benchmark{} (Long-Horizon \underline{\vphantom{g}M}emory under \underline{\vphantom{g}INT}erference \underline{\vphantom{g}Eval}uation), which features \emph{interference-heavy input contexts}, queries requiring \emph{long-range lookback and aggregated reasoning}, as well as \emph{diverse domain and question types}. 
As shown in Figure~\ref{fig:fig1} left, \benchmark spans \textit{four} domains~(state tracking, multi-turn dialogue, Wiki revisions, and Git commits), each involving continuously evolving information streams with accumulated context. The evolution covers both overwrite-style (edit-based) and append-style (accumulative) streams, enabling evaluation across different memory dynamics under interference-heavy scenarios.
The benchmark also includes \textit{two} primary types of tasks\footnote{More examples for each question type are in Table~\ref{table: Q_example}.}: \textit{Single-target recall} tasks evaluate whether models can accurately retrieve specific pieces of information under interference; (e.g., \emph{``According to the previous revision of the article, how many floors does the building have?''}).
\textit{Multi-target aggregation} tasks require models to identify and perform \emph{aggregated reasoning} over multiple relevant pieces of context, including operations such as counting entities, ordering events, and combining information across updates. 
For example, a multi-target query like 
\emph{``What syntax changes were made between version 1.2.30 and the current package versions?''} requires recalling the syntax of both version 1.2.30 and the current version, and then reasoning over the differences between them.
We construct \benchmark using both synthetic examples from existing benchmarks and LLM-generated questions produced by Gemini-3.1-Pro~\citep{gemini31pro} conditioned on the full interaction history. 
Overall, \benchmark is a diverse and scalable benchmark containing an average of 3.9k questions per domain and 15.6k question-answering pairs in total, built over long-horizon contexts averaging 138.8k tokens and extending up to 1.8M tokens.
Each context contains, on average, 86 temporally ordered updates.
For questions that are generated by the frontier model, we further conduct a human verification with six annotators on 20\% instances and find that 95.6\% of them are valid.

Using \benchmark, we evaluate seven representative systems using \qwen~\citep{qwen3technicalreport} and \gemini~\citep{gemini31flash}: Full Context, RAG, HippoRAG~\citep{gutierrez2025ragmemorynonparametriccontinual}, MemAgent~\citep{yu2025memagentreshapinglongcontextllm}, AtomMem~\citep{huo2026atommemlearnabledynamic}, Mem-$\alpha$~\citep{wang2025memalpha}, and SimpleMem~\citep{liu2026simplememefficientlifelongmemory}. 
Across all systems, \benchmark remains highly challenging, with an average accuracy of 27.9\%; the best-performing system, MemAgent, achieves only 33.4\% on average, with failure modes described in Fig. \ref{fig:fig1} (right). 
We observe that performance varies across tasks and domains. 
In particular, memory management systems perform strongly on \babi~\citep{weston2015towards}, which contains relatively short contexts and simple facts, achieving an average improvement of +9.9\% over non-memory baselines. However, on other domains with longer contexts and evolving revisions, these systems often underperform the same baselines, with an average 3.0\% drop.
Also, performance differs significantly by question type: simple recall questions have higher average accuracy (47.5\%), whereas systems perform poorly on questions requiring long-range lookback (avg. 21.0\%), and those requiring \textit{multi-target aggregation} (avg. 26.5\%).
To better understand where these failures occur, we decompose errors into (1) failures in retrieval or memory construction, and (2) failures of the answering agent to correctly use relevant information even when it is available in the context. Our analysis shows that most errors stem from memory construction failures, which account for a 41.7\% performance drop, while the answering stage contributes an additional 25.2\% drop.
Further analysis shows that memory-augmented agents are sensitive to design choices such as the number of iterative memory process steps, and are strongly biased toward insertion-based operations (avg. 76.8\%) instead of deletion or update.
Overall, our analysis reveals key strengths and limitations of existing memory systems, emphasizing the need for approaches that are robust to interference-heavy contexts, domain generalization, and various queries, including long-range lookback and aggregated reasoning.

\section{\benchmark: Long-Horizon \underline{\vphantom{g}M}emory under \underline{\vphantom{g}INT}erference \underline{\vphantom{g}Eval}uation}

\paragraph{Interference-heavy Contexts.}
\label{sec3: input_context}

\benchmark focuses on contexts with \textit{densely interacting updates}, where information is repeatedly modified or contradicted over time~(Figure~\ref{fig:fig1}, middle). Real-world memory involves continual revisions and conflicting states. These dynamics expose the core challenges of memory systems: resolving temporal conflicts, preserving historical state, and maintaining consistency over time. Such setups naturally induce \emph{proactive and retroactive interference}~\citep{underwood1957interference, ANDERSON1996237} where retroactive interference occurs when new information disrupts recall of older information, while proactive interference occurs when older memories interfere with learning or recalling newer information. By incorporating both, our setup requires agents to track evolving states, connect historical information, and resolve interference effectively.

\paragraph{Domains.} \benchmark consists of four representative domains in which memory is frequently helpful in practice.
These domains differ in information structure, update dynamics, and reasoning requirements, enabling evaluation of both memory behavior under varied interference patterns and \emph{domain generalization} across tasks (Examples and more details are in Appendix~\ref{app_benchmark: domains}.).

\textsc{(1) State Tracking} (\babi). 
We use contexts from \babi~\citep{weston2015towards}, where information is represented as simple symbolic facts that are updated through sequential, localized changes, often overwriting previous states.
Questions query the changing states and facts described in the context. 
This domain requires systems to integrate sequential updates, track state transitions, and perform temporal reasoning over current and historical states. 

\textsc{(2) Dialogue-based Multi-turn Interactions} (\horizonbench).
Building on \horizonbench~\citep{li2026horizonbench}, a long-horizon personalization benchmark with users and conversation histories, we form long-horizon multi-turn dialogue contexts by concatenating multiple dialogue sessions. 
We then generate new questions targeting personal preferences and attributes whose relevant information is distributed across interactions and often implicitly expressed through natural language interactions. 
This domain evaluates whether memory systems can track and update \textit{implicit} user-state changes, such as evolving preferences, over time.

\textsc{(3) Factual Knowledge QA} (Wiki Revisions).
We introduce the Wiki Revisions split, which we construct from long-horizon Wikipedia revision histories, where each instance consists of chronologically ordered article revisions. 
We generate questions targeting both factual knowledge in the articles and how information evolves across revisions. 
As facts may be added, modified, contradicted, or removed over revisions, answering these questions requires memory systems to reconstruct prior states, track provenance, and distinguish outdated from current information.

\textsc{(4) Code and Files Evolution} (Git Commits).
We also introduce the Git Commits splits, which constructs long-horizon contexts from GitHub commit histories, where each instance contains a single repository and its chronological commits. We construct questions that target both code details in the repository and how implementations evolve across commits. Unlike natural-language revision histories, code evolution often involves tightly coupled cross-file edits and evolving identifiers (e.g., function name or API signature), thus requiring a memory system to recover implicit differences between snapshots and changing program behavior.

\begin{table*}[t!]
\centering
\small
\caption{
Example from each question type in \benchmark. 
}
\fontsize{7}{10} 
\begin{tabular}{c|c}
\toprule
\multicolumn{2}{c}{Single-Target Recall} \\
\midrule
Simple & How many floors does the article state the building has? \\
History & In the version two edits prior, which team is named 1919 County Champion? \\
\midrule
\multicolumn{2}{c}{Multi-Target Aggregation} \\
\midrule
Ordering & In which order was the section added to the article? \\
Counting & How many different individuals have been listed as the album’s producer? \\
Multihop & What was episode 5’s title just before episode 4’s third title change? \\
\bottomrule
\end{tabular}
\label{table: Q_example}
\end{table*}

\paragraph{Question Types.} 
\label{sec3: qtypes}

\benchmark includes two primary categories of tasks that target different aspects of memory behavior under densely interacting updates and interference-heavy contexts: \textsc{single-target recall} and \textsc{multi-target aggregation} (Examples in Table~\ref{table: Q_example}).

\textsc{Single-Target Recall.}
These tasks evaluate whether a model can correctly identify and retrieve a single target from long contexts with dense updates. 
We consider \textit{two} variants: \simple questions, which require retrieving the most recent state after a sequence of updates, and \history (\textit{lookback-style}) questions, which require recovering an earlier state despite subsequent updates and potentially conflicting information.
\simple questions evaluate robustness to \textit{proactive} interference, where previously stored information may interfere with encoding or retrieving newer states. In contrast, \history questions evaluate robustness to \textit{retroactive} interference, where newly introduced information may overwrite or obscure previously stored states. \history questions require agents to identify the relevant point in the context using cues and respond using the corresponding information.
Together, these tasks evaluate whether models can both maintain up-to-date representation and preserve access to prior states over long contexts. 

\textsc{Multi-Target Aggregation.}
These tasks require agents to identify multiple targets distributed across different updates and aggregate them to produce the correct answer. 
We consider \textit{three} variants based on the type of aggregation required.
(1) \order questions require recovering the correct temporal order of events under dense updates.
(2) \counting questions require aggregating occurrences across updates, such as determining how many times an event happened or how long a particular state persisted.
(3) \multihop questions require reasoning over multiple targets, such as comparing information across updates or performing bridge reasoning over interdependent events. 
These three tasks evaluate whether models can identify multiple targets, integrate information across updates, and reason over their relationships despite interference from intervening updates.

\textbf{Question Generation Pipeline.}
Depending on the availability and structure of metadata in each domain, we adopted different procedures for constructing question-answer pairs. 
For \babi, we parsed each fact into a (subject, object, verb) tuple and generated a question by filling predefined templates with the extracted information, following a procedure similar to \cite{kim2026largelanguagemodelsup}. 
For \horizonbench, we used the metadata provided by \cite{li2026horizonbench}, which tracks temporal changes such as evolving user preferences. We constructed question templates and filled them using the metadata, similar to \babi.
For Wiki Revisions and Git Commits, we generate question-answer pairs by prompting Gemini-3.1-Pro~\citep{gemini31pro} with revision metadata, including \texttt{revision\_ids}, \texttt{timestamp}, \texttt{editor}, \texttt{comment}. 
We conduct a human validation process with six annotators, including three authors and three non-authors, on 20\% of the sessions (40 out of 200 sessions for Git Commits and 42 out of 196 sessions for Wiki Revisions). For each session, annotators are asked to evaluate one question-answer pair from each question type, for question naturalness and answer correctness. The results show that 95.6\% of the generated samples contain natural questions with correct answers.
More details about question generation and human validation are in Appendix~\ref{app_benchmark: q_generation}.

\textbf{Dataset Statistics.}
Table~\ref{table: Q_stats_simple_ver} summarizes the scale and composition of \benchmark across domains. 
On average, each domain contains 149 sessions, with contexts averaging 86 updates in depth and 138.8k tokens in length.
Across domains, \benchmark includes an average of 2k questions for \textit{single-target recall} and 1.8k for \textit{multi-target aggregation}. More details are in Appendix~\ref{app_benchmark: stats}.
\begin{table*}[t!]
\centering
\small
\caption{
Dataset statistics across four domains.
\textit{Depth} denotes the number of turns, revisions, or commits in each example.
\textit{k} indicates values reported in thousands. Further details in Appendix~\ref{app_benchmark: stats}.
}
\begin{tabular}{cl|cccc}
\toprule
 && \babi & HorizonBench & Wiki Revisions & Git Commits \\
\midrule
&\# Sessions & 99 & 100 & 196 & 200 \\
&\# Total Questions & 5.7k & 6.9k & 1.5k & 1.6k \\
\midrule
\multirow{2}{*}{\shortstack{\textit{Avg. Context}\\\textit{Statistics}}}
&Depth & 42  & 142 & 99  & 61 \\
&Tokens & 0.3k  & 274k  & 195k  & 86k \\

\midrule
\multirow{2}{*}{\shortstack{\textit{Question}\\\textit{Distributions}}}
&Single-Target Recall & 2.7k  & 3.9k  & 0.8k & 0.9k \\
&Multi-Target Agg. & 3k & 3k & 0.6k & 0.7k \\
\bottomrule
\end{tabular}
\label{table: Q_stats_simple_ver}
\end{table*}

\section{Experiments}
\subsection{Setup}
\noindent\textbf{Baselines.}
Our baselines fall into three main categories.
\textbf{(1) Full Context:} methods without an explicit memory module, where the model receives the entire context as input.
\textbf{(2) Retrieval-Augmented Generation (RAG):} \textsc{RAG} denotes the standard retrieval-augmented generation framework, which retrieves relevant documents using dense vector similarity~\citep{lewis2020retrieval}. \textsc{HippoRAG}~\citep{gutierrez2025ragmemorynonparametriccontinual} extends this framework with a graph-structured retrieval mechanism that captures richer relationships between documents. Unless otherwise specified, we retrieve the top-5 contexts.\footnote{We provide an analysis of performance under different numbers of retrieved documents in Appendix~\ref{app: RAG}, where we observe that retrieving five documents provides a strong overall performance.}
\textbf{(3) Memory-Augmented Agents:} We evaluate several trained memory systems that explicitly learn how to store, update, and retrieve information under different training paradigms. For all methods, we use the officially released checkpoints. For \babi, every 15 facts are grouped into a single chunk. For \horizonbench, each dialogue session is treated as a chunk; for Wiki Revisions and Git Commit, each revision is treated as a chunk.\footnote{We additionally provide an ablation study on chunk size in Section~\ref{sec4: memory_agents_ablation}.}  
\textsc{MemAgent}~\citep{yu2025memagentreshapinglongcontextllm} is built on Qwen2.5-14B-Instruct~\citep{qwen2.5}, and it incrementally updates memory using an overwriting strategy, constructing query-specific memory representations. 
\textsc{AtomMem}~\citep{huo2026atommemlearnabledynamic} formulates memory management as a sequential decision-making problem, decomposing actions into atomic CRUD (Create, Read, Update, Delete) operations, and is based on Qwen3-8B~\citep{qwen3technicalreport}.
\textsc{Mem-$\alpha$}~\citep{wang2025memalpha} trains Qwen3-4B model to organize memory into three types, i.e., core, semantic, and episodic memory. 
\textsc{SimpleMem}~\citep{liu2026simplememefficientlifelongmemory} is a state-of-the-art memory system consisting of a three-stage pipeline: semantic structured compression, which converts unstructured interactions into compact multi-view memory units; online semantic synthesis, which incrementally merges related contexts to reduce redundancy; intent-aware retrieval, which dynamically determines retrieval scope and constructs targeted retrieval contexts. 

\noindent\textbf{Models.}
Our evaluation pipeline consists of three components. (1) \textit{Memory manager} constructs a compact memory representation of a long-horizon, evolving input context. For \textsc{SimpleMem}, we use \gemini~\citep{gemini31flash}. For \textsc{Mem-$\alpha$}, \textsc{MemAgent}, and \textsc{AtomMem}, we use their publicly released checkpoints.
(2) \textit{Answering agent} takes either the full context, retrieved context, or managed memory as input and generates the final answer. 
Unless otherwise specified, we use \qwen~\citep{qwen3technicalreport} as the answering agent and additionally evaluate \gemini. We set the maximum context length to 65k and 1M tokens for \qwen and \gemini, respectively. 
(3) \textit{Embedding model} is used in retrieval-based systems to retrieve relevant contexts by computing similarity scores. 
Unless otherwise specified, we use Qwen3-Embedding-4B~\citep{qwen3embedding} and additionally evaluate Gemini-Embedding-001~\citep{geminiemb}. 
Further details are provided in Appendix~\ref{app_exp: setup}.

\noindent\textbf{Evaluation Metrics.}
We evaluate using Exact Match after standard text normalization, following prior memory benchmarks~\citep{kim2026largelanguagemodelsup, wang2025memalpha}.
For \horizonbench only, we provide a set of candidate answers for each question, similar to a multiple-choice evaluation setting, since answers may not appear verbatim in the context and can admit multiple valid surface forms. 

\begin{table*}[t!]
\centering
\small
\caption{
Results on \qwen. We compare three categories of methods: \textcolor{FullBaseline}{Full Context}, \textcolor{RAGBaseline}{RAG}, and \textcolor{MemBaseline}{Memory-Augmented Agents}. 
Cells are color-coded by score to highlight performance patterns, transitioning from dark red (lowest) through light red and light green to dark green (highest).
} 
\fontsize{7}{10} 
\begin{tabular}{cccccccc}
\toprule
 & & \textcolor{FullBaseline}{Full} & \textcolor{RAGBaseline}{RAG} & \textcolor{RAGBaseline}{HippoRAG} & \textcolor{MemBaseline}{AtomMem} & \textcolor{MemBaseline}{Mem-$\alpha$} & \textcolor{MemBaseline}{MemAgent}  \\
\midrule
\multirow{5}{*}{\babi } 
& Simple   & \cv{57.4} & \cv{66.7} & \cv{70.0} & \cv{65.2} & \cv{82.6} & \cv{85.7} \\
& History  & \cv{16.1} & \cv{16.7} & \cv{33.3} & \cv{36.3} & \cv{44.9} & \cv{36.0} \\
& Ordering & \cv{22.0} & \cv{37.5} & \cv{50.0} & \cv{58.1} & \cv{64.7} & \cv{59.0} \\
& Counting & \cv{40.5} & \cv{80.0} & \cv{80.0} & \cv{43.8} & \cv{70.4} & \cv{24.3} \\
& Multihop & \cv{30.8} & \cv{40.0} & \cv{41.7} & \cv{30.7} & \cv{61.0} & \cv{51.7} \\
\midrule
\multirow{5}{*}{Wiki Revisions} 
& Simple   & \cv{23.3} & \cv{36.7} & \cv{37.9} & \cv{16.9} & \cv{49.9} & \cv{54.2} \\
& History  & \cv{14.5} & \cv{30.2} & \cv{31.1} & \cv{15.7} & \cv{20.6} & \cv{28.8} \\
& Ordering & \cv{10.9} & \cv{6.9}  & \cv{4.1}  & \cv{2.4}  & \cv{13.5} & \cv{38.3} \\
& Counting & \cv{22.2} & \cv{15.9} & \cv{19.1} & \cv{14.3} & \cv{25.0} & \cv{36.5} \\
& Multihop & \cv{11.2} & \cv{26.8} & \cv{27.1} & \cv{16.2} & \cv{17.3} & \cv{23.7} \\
\midrule
\multirow{5}{*}{Git Commits} 
& Simple   & \cv{82.0} & \cv{81.5} & \cv{81.9} & \cv{40.8} & \cv{71.7} & \cv{82.3} \\
& History  & \cv{27.1} & \cv{30.1} & \cv{30.6} & \cv{19.0} & \cv{4.8}  & \cv{24.0} \\
& Ordering & \cv{17.7} & \cv{40.6} & \cv{44.5} & \cv{13.8} & \cv{17.0} & \cv{55.9} \\
& Counting & \cv{21.1} & \cv{38.9} & \cv{14.1} & \cv{24.7} & \cv{18.8} & \cv{51.6} \\
& Multihop & \cv{13.1} & \cv{12.8} & \cv{39.6} & \cv{27.3} & \cv{8.5}  & \cv{34.7} \\
\midrule
\multirow{5}{*}{HorizonBench} 
& Simple   & \cv{11.3} & \cv{11.6} & \cv{12.1} & \cv{4.4}  & \cv{7.5}  & \cv{7.5}  \\
& History  & \cv{9.9}  & \cv{10.3} & \cv{10.9} & \cv{2.4}  & \cv{5.8}  & \cv{3.8}  \\
& Ordering & \cv{3.5}  & \cv{4.2}  & \cv{3.9}  & \cv{1.0}  & \cv{0.4}  & \cv{6.7}  \\
& Counting & \cv{0.8}  & \cv{2.8}  & \cv{4.2}  & \cv{2.9}  & \cv{1.6}  & \cv{1.8}  \\
& Multihop & \cv{11.9} & \cv{29.7} & \cv{33.9} & \cv{30.0} & \cv{24.0} & \cv{28.1} \\
\midrule
Overall Avg. & & \cv{21.0} & \cv{29.5} & \cv{32.3} & \cv{22.1} & \cv{28.0} & \cv{33.4} \\
\bottomrule
\end{tabular}
\label{table: main_results}
\end{table*}

\subsection{Results}

\paragraph{Existing Methods Struggle on \benchmark{}.}
As shown in Table~\ref{table: main_results}, existing systems struggle on \benchmark, achieving only 27.7\% average accuracy across the six evaluated systems.
Even advanced memory systems perform poorly: the best overall result reaches just 33.4\% averaged across all domains, suggesting that the benchmark remains far from saturated.
Across question types, both RAG and memory-based methods perform relatively well on \simple queries (avg. 47.5\%), suggesting that retrieving the most recent value is comparatively easy. However, performance drops substantially on \history questions that require long-range lookback (avg. 21.0\%) and on \textit{multi-target aggregation} questions (avg. 26.5\%), which require tracking updates over time, resolving conflicts, or aggregating information across multiple targets.
Among memory-based approaches, MemAgent achieves the strongest overall performance (avg. 33.4\%) and shows relatively robust generalization across domains. We hypothesize that this gain comes from the construction of query-specific memory representations, whereas AtomMem and Mem-$\alpha$ build a shared memory from the input context and reason over the same question-agnostic memory structure.  
Nevertheless, MemAgent's average performance on \benchmark remains low, indicating that \benchmark is challenging even for strong existing memory systems. 
Further analysis of online evaluation performance is in Appendix~\ref{app: online_eval}, and an analysis of performance with partial credit is in Appendix~\ref{app: partial_correctness}.

\paragraph{\benchmark Shows Limited Cross-Domain Generalization.}
The overall results in Table~\ref{table: main_results} exhibit substantial variance, with no single method consistently outperforming others across domains and question types. For example, MemAgent achieves 85.7\% on \babi \simple but drops to 7.5\% on \horizonbench for the same task, while HippoRAG attains 70.0\% on \babi \simple and remains relatively more robust on \horizonbench \simple with 12.1\%. These results suggest limited cross-domain generalization.
In general, single-target recall tasks (\simple and \history) are easier than multi-target aggregation tasks (\order, \counting, and \multihop), with average accuracies of 34.3\% and 26.5\%, respectively. This gap arises because aggregation tasks require identifying multiple relevant targets and performing additional reasoning over them. Within single-target recall, \history questions (21.0\%) are consistently harder than \simple questions (47.5\%) since they require retrieving past rather than current states, with difficulty increasing for longer lookback distances (Section~\ref{sec4: lookback}).
Among aggregation tasks, \order questions are most difficult (24.0\%) as they require recovering the exact event sequence without partial credit. 
Overall, \benchmark highlights persistent challenges with interference-heavy contexts and long-horizon dependencies, with large performance gaps across both domains and question types.

\paragraph{Even the State-of-the-art Memory System Struggles.}
We further evaluate SimpleMem~\citep{liu2026simplememefficientlifelongmemory}, a state-of-the-art memory system using frontier models (\gemini as answering agent and Gemini-Embedding-001 as embedding model), to investigate the performance of a strong memory system combined with the frontier models. 
Despite using a stronger embedding model and answering agent, SimpleMem achieves only 30.3 EM on average.
We find that this degradation stems from SimpleMem's aggressive memory compression strategy. Such compression is effective on conversational memory benchmarks such as LoCoMo~\citep{maharana2024evaluating}, where contexts are relatively short (avg. 109 characters) and less interconnected. In contrast, as \benchmark contains long, evolving revisions (avg. 184k characters) with substantial interdependence and interference, and thus, aggressive compression and deduplication are prone to discarding important provenance information and historical details.
Consistent with the observation, SimpleMem performs relatively well on \babi, which contains shorter and simpler contexts, but degrades substantially on Wiki Revisions. In particular, revision provenance is often lost during compression, as facts may be paraphrased or rewritten. Without explicit metadata linking facts to their originating revisions, retrieval relies primarily on keywords and embeddings, making queries such as retrieving the content of ``Revision 53'' especially challenging. Further analyses are provided in Appendix~\ref{app_exp: simplemem}.

\section{Analysis}

\begin{figure}[t!]
    \centering
    \begin{minipage}[t]{0.49\linewidth}
    \centering
    \includegraphics[width=\linewidth]{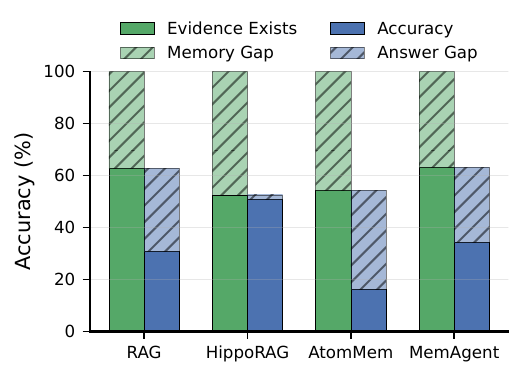}
    \vspace{-14pt}
    \caption{
    Error due to missing evidence in memory (green) or incorrect answers despite the evidence being present (green–blue gap).
    Only 58.3\% of cases contain the required evidence, making retrieval/memory construction the main bottleneck; answering errors add a 25.2\% drop. A perfect system would reach 100\%.
    }
    \label{fig: evidence_exists}
    \end{minipage}
    \hfill
    \begin{minipage}[t]{0.49\linewidth}
    \centering
    \includegraphics[width=\linewidth]{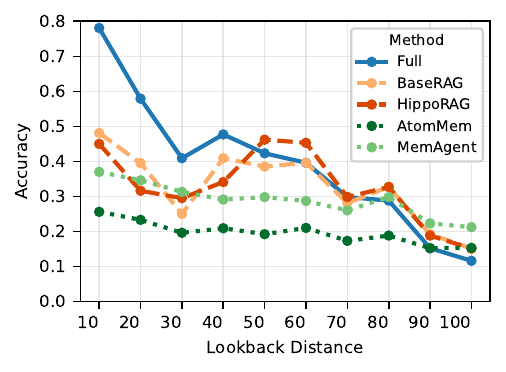}
    \vspace{-14pt}
    \caption{
    Performance (y-axis) vs. Lookback Distance (x-axis). Accuracy drops as lookback distance increases, with the largest drops in Full Context and retrieval methods (RAG, HippoRAG). Memory-based agents degrade less, suggesting greater robustness from better temporal encoding and compact historical representations.
    } 
    \label{fig: n_step_back}
    \end{minipage}
\end{figure}

\subsection{Retrieval and Memory Construction Remain the Primary Bottleneck}
\label{sec4: retrieval_memory}

We note that the RAG and memory-based systems consist of two stages: (1) retrieving relevant information or constructing memory, and (2) generating an answer conditioned on the retrieved context or constructed memory using an answering agent. 
Failures can therefore arise from two sources: failures in retrieval or memory construction, or in answer generation.
To investigate the source of these failures, we analyze two retrieval systems (RAG and HippoRAG) and two memory systems (MemAgent and AtomMem) on the Wiki Revisions.
Using \gemini, we determine whether failures arise from retrieval/memory construction or answer generation by checking whether the retrieved documents or constructed memories contain the supporting evidence required to answer the question.\footnote{We use an LLM-based evaluation for analysis instead of lexical matching because we observe that the same words may appear multiple times in the context without being relevant to the question, making simple word matching imprecise.} We analyze on \simple, \history, and \multihop questions, as answers to \order and \counting questions are often not explicitly stated in the retrieved context. 

In ~\cref{fig: evidence_exists}, we view 100\% as the upper bound, since all questions are generated directly from the full history, meaning that the required evidence always exists by design in the retrieval pool. 
Relative to this upper bound, the largest performance degradation comes from retrieval and memory construction failures, resulting in an average drop of 41.7\% (only 58.3\% of cases contain the supporting evidence). 
When the evidence is present, an average of 25.2\% drop can be attributed to failures of the answering agent (blue bars). 
These findings indicate that current retrieval and memory construction are the primary bottleneck, while the strength of the answering agent also plays a non-trivial role in performance.
Although all four systems use the same answering agent, differences in how retrieved information and memories are constructed and presented lead to substantial performance gaps. 
For example, AtomMem shows particularly large degradation in answer generation performance, as we observe that it produces relatively longer memories compared to other methods.
We further analyze the effect of different answering agents on performance in Appendix~\ref{app: memagent_answering_agent}.
Under the Full Context setting, replacing answering agent from \qwen to \gemini yields a substantial performance improvement (55.7\%). In contrast, this gap becomes much smaller when retrieval or memory systems are introduced (avg.~1.7\%), indicating that once memory systems are involved, performance differences are driven less by the capability of the answering agent itself and more by how effectively the retrieval or memory system constructs context for the agent.

\subsection{Longer Lookback Distances Hurt Performance, and Temporal Markers Help}
\label{sec4: lookback}
We analyze how performance changes as the required lookback distance increases for \history questions, which ask about information from earlier revisions (e.g., `In the version \textit{two edits prior}, which team is named 1919 County Champion?'). 
Here, lookback distance refers to the number of revisions between the queried information and the current version. 
We evaluate five settings on the Wiki Revisions subset across questions with varying lookback distances.
As shown in Figure~\ref{fig: n_step_back}, performance generally decreases as the required lookback distance increases, suggesting that retrieving or preserving information from distant revisions is increasingly difficult.
The largest degradation is observed for the Full Context setup and retrieval-based methods (RAG and HippoRAG), whose accuracy drops substantially as the number of lookback distance grows. In contrast, although memory-augmented agents also exhibit some degradation, the decline is noticeably smaller. We hypothesize that this greater robustness arises because memory-based agents can better encode temporal order and preserve relationships between events by accumulating historical information into memory.
We further investigate how incorporating explicit temporal cues (e.g., October 2023) into the context and questions affects performance. 
As shown in Appendix~\ref{app_exp: adding_date_and_time}, we observe that explicit temporal cues substantially reduce performance drops, suggesting that they mitigate interference by helping agents distinguish similar or conflicting facts across revisions.

\begin{figure}[t!]
    \centering
    \begin{minipage}[t]{0.48\linewidth}
    \centering
    \includegraphics[width=\linewidth]{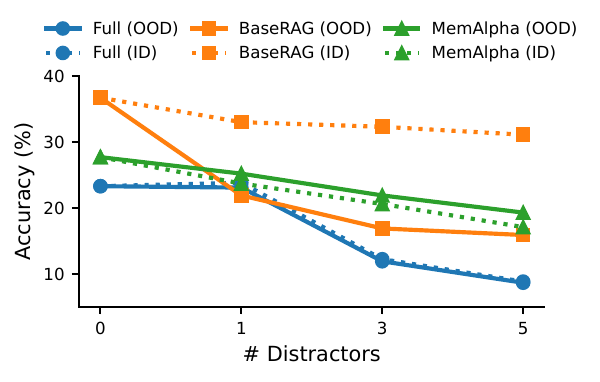}
    \vspace{-14pt}
    \caption{Performance under varying numbers of distractors for both \textit{In-Domain (ID)} and \textit{Out-of-Domain (OOD)} settings. 
    Overall, the performance drops as the number of distractors increases, while Full Context shows no significant difference across \textit{ID} and \textit{OOD} distractors. 
    }
    \label{fig: distractor_type_num}
    \end{minipage}
    \hfill
    \begin{minipage}[t]{0.48\linewidth}
    \centering
    \includegraphics[width=\linewidth]{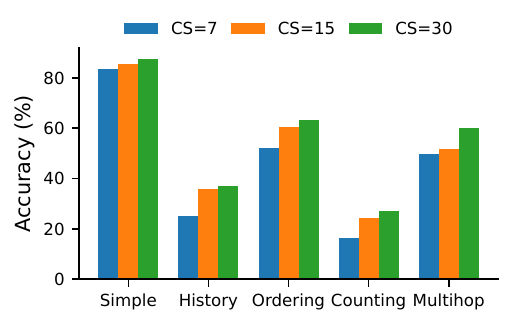}
    \vspace{-14pt}
    \caption{
    Performance vs. different chunk sizes when processing memories for the MemAgent model (CS = Chunk Size). 
    Increasing CS generally improves performance, and \simple questions are the least sensitive to CS, since it only requires recalling recent information.
    } 
    \label{fig: memory_chunk_size}
    \end{minipage}
\end{figure}

\subsection{Adding Distractors Further Degrades Performance, Especially for RAG}
To evaluate how agents perform in real-world scenarios with noisy distractors, we study how performance changes as different types and numbers of distractors are inserted between facts in the \babi dataset.
We insert two types of sentence-level distractors with varying numbers of inserted sentences (1, 3, and 5), measuring performance across Full, RAG, and Mem-$\alpha$.\footnote{In both cases, we ensure distractors do not alter the answer by removing sentences sharing the same subject and object.}
The distractor types are: (1) \textit{Out-of-Domain (OOD)} distractors drawn from novels (similar to BABILong~\citep{kuratov2024babilong}), which differ in style and structure from \babi facts; 
(2) \textit{In-Domain (ID)} distractors, which follow the same simple, compositional structure as \babi facts. 
As shown in ~\cref{fig: distractor_type_num}, performance generally decreases as the number of distractors increases across all agents. 
However, the degradation is most pronounced with \textit{OOD} distractors for RAG, which tends to retrieve distracting sentences more frequently. 
In contrast, for Mem-$\alpha$ and the Full Context baseline, the difference in performance between \textit{OOD} and \textit{ID} distractors is relatively small. 
We provide a more fine-grained analysis in \cref{app_fig: distracting} in the Appendix.

\subsection{Ablation Studies and Analysis of Memory-Augmented Agents}
\label{sec4: memory_agents_ablation}
\noindent\textbf{Fewer Memory Update Iterations Improve Performance.}
In memory systems, long contexts can be processed using different chunk granularity (e.g., a 1M token input can be divided into 10 100k-sized chunks or 100 10k chunks). We investigate how different chunk sizes, which determine the number of memory update iterations, affect overall performance on \babi using MemAgent. 
As shown in Figure~\ref{fig: memory_chunk_size}, increasing the chunk size, i.e., reducing the number of memory modifications, generally improves performance as more frequent modifications may introduce unintended overwrites or removals of previously stored information.
This is especially apparent for \history and \counting questions, which require integrating information across long horizons, and less for \simple questions, which typically rely on recent information.

\paragraph{Existing Memory Systems Strongly Biased Toward Appending Rather than Editing or Deleting.}
Both AtomMem and Mem-$\alpha$ manage memory through function calls corresponding to three operations: (1) insertion, (2) modification, and (3) deletion.
Analyzing the frequency of these operations, we observe that both systems are heavily biased toward insertion across all datasets, which accounts for 87.6\% of operations in AtomMem and 65.9\% in Mem-$\alpha$ on average (Figure~\ref{app_fig: tool_usage_rate} in Appendix).
This suggests that, although revisions in \benchmark are often incremental refinements of earlier memory entries, agents struggle to recognize these updates as modifications because many changes are frequently expressed implicitly and relationships between revisions are not properly captured, leading to redundant memory insertions. 
This issue is further exacerbated by the coarse granularity of memory operations. 
Both systems tend to operate on large chunks rather than in fine-grained units, making it difficult to detect and update small differences within existing entries. 
As a result, even minor changes are often inserted as new information instead of modifying existing memory. 
Overall, these findings highlight the need for more balanced memory management.
Detailed results are in Appendix~\ref{app: tool_usage_rate} and an analysis of \benchmark as training set in Appendix~\ref{app: train}.\footnote{We further conduct ablation studies and analysis over RAG performance in Appendix~\ref{app: RAG}.}

\section{Related Work}
Various memory-augmented agents have been developed including pipeline-based systems (MemGPT~\citep{packer2024memgptllmsoperatingsystems} and SimpleMem~\citep{liu2026simplememefficientlifelongmemory}); training-based approaches (MemAgent~\citep{yu2025memagentreshapinglongcontextllm} and MEM1~\citep{zhou2025mem1learningsynergizememory}, Memory-R1~\citep{yan2025memory}, and AtomMem~\citep {huo2026atommemlearnabledynamic}); and architectures with different memory types (e.g., episodic, semantic memory) (Mem-$\alpha$~\citep{wang2025memalpha} and REMem~\citep{shu2026rememreasoningepisodicmemory}, and SYNAPSE~\citep{jiang2026synapseempoweringllmagents}). 
To evaluate these systems, several benchmarks have been proposed that measure memory capabilities along different dimensions, including conversation benchmarks~\citep{maharana2024evaluating, wu2025longmemevalbenchmarkingchatassistants}, QA-based benchmarks~\citep{hu2026evaluatingmemoryllmagents}, and more recently, benchmarks using densely interconnected contexts have been released, such as OAKS~\citep{kim2026largelanguagemodelsup}, StoryBench~\citep{wan2025storybenchdynamicbenchmarkevaluating} and RealMem~\citep{bian2026realmembenchmarkingllmsrealworld}. 
However, as shown in Table~\ref{table: comparison}, existing benchmarks remain limited in three key aspects: (1) lack interference-heavy contexts, (2) limited evaluation of domain generalization, and (3) cover few question types. \benchmark addresses these limitations, enabling evaluation of memory systems in more realistic settings. More details in Appendix~\ref{app: related_work}.

\section{Conclusion}
We introduce \benchmark, an analytical benchmark characterized by interference-heavy contexts, long-range dependencies, and multi-target aggregation reasoning. 
It spans four domains (state tracking, multi-turn dialogue, Wikipedia revisions, and GitHub commits) and five question types covering both single-target recall and multi-target aggregation.
\benchmark remains far from saturated: the average accuracy across systems is only 27.9\%, and the strongest model achieves just 33.4\%. 
We observe that performance degrades substantially on questions that require lookback or aggregated reasoning, with retrieval and memory construction emerging as the dominant bottleneck. 

\section*{Acknowledgments}
We would like to thank the annotators: Hanqi Xiao, Vu Hoang Thien An, and Jefrey Bergl. This work was supported by Microsoft Agentic AI Research and Innovation (AARI) grant program, NDSEG PhD Fellowship, NSF-AI Engage Institute DRL-2112635, and NSF-CAREER Award 1846185. The views contained in this article are those of the authors and not of
the funding agency.

\bibliographystyle{plainnat}
\bibliography{ref}

\appendix

\newpage
\appendix

\section{Additional Benchmark Details}

\subsection{Additional comparison with existing benchmarks}
\label{app:more_benchmarks}

\begin{table*}[t!]
\centering
\small
\caption{
Comparison of \benchmark with prior memory benchmarks. We categorize benchmarks by (1) input context properties: interdependent inputs (\textbf{Interdep.}), dense interference ($\geq$10 depth; \textbf{Interference}), and multi-domain coverage (\textbf{M-Domain}); and (2) question properties: multi-target aggregation (\textbf{Aggr.}) and lookback to earlier context (\textbf{LookBack}).
}
\begin{tabular}{lccc|cc}
\toprule
& \multicolumn{3}{c}{Input Context} & \multicolumn{2}{c}{Question Type} \\
 \cmidrule{2-4} \cmidrule{5-6}
Benchmark & Interdep. & Interference & M-Domain & Aggr.& LookBack \\
\midrule 
HotpotQA~\citep{yang2018hotpotqadatasetdiverseexplainable}  & \greencmark & \redxmark & \redxmark  & \greencmark & \redxmark \\
MuSiQue~\citep{trivedi2022musiquemultihopquestionssinglehop}  & \greencmark & \redxmark & \redxmark  & \greencmark & \redxmark \\
2WikiMultiHopQA~\citep{ho2020constructingmultihopqadataset}  & \greencmark & \redxmark & \redxmark  & \greencmark & \redxmark \\
RULER~\citep{hsieh2024rulerwhatsrealcontext}  & \redxmark& \redxmark & \redxmark  & \greencmark & \redxmark \\
HELMET~\citep{yen2025helmetevaluatelongcontextlanguage}  & \greencmark& \redxmark & \greencmark  & \greencmark & \redxmark \\
NarrativeQA~\citep{kočiský2017narrativeqareadingcomprehensionchallenge}  & \greencmark& \redxmark & \redxmark  & \greencmark & \redxmark \\
lost in conversation~\citep{handa2026lostconversationlosttranslation}  & \greencmark& \redxmark & \redxmark  & \greencmark & \redxmark \\
$\tau$-bench~\citep{yao2024taubenchbenchmarktoolagentuserinteraction}  & \greencmark& \redxmark & \redxmark  & \greencmark & \redxmark \\
SWE-bench~\citep{jimenez2024swebenchlanguagemodelsresolve}  & \greencmark& \redxmark & \redxmark  & \greencmark & \redxmark \\
\rowcolor{blue!7}\benchmark (Ours) & \greencmark & \greencmark & \greencmark & \greencmark & \greencmark \\
\bottomrule
\end{tabular}
\label{table: add_comparison}
\end{table*}

We expand Table~\ref{table: comparison} to more existing benchmarks in Table~\ref{table: add_comparison}.
While the previous benchmarks in Table~\ref{table: add_comparison} may need memory-related capabilities as part of end-to-end task success, they primarily evaluate overall task performance rather than the memory management process itself. 
In contrast, \benchmark focuses on whether a memory system can construct, maintain, and utilize coherent memory representations as related observations continuously accumulate over time. 
Rather than simply retrieving evidence from long contexts, \benchmark requires memory systems to maintain dependencies across observations, handle interference introduced by subsequent information, aggregate evidence across multiple updates, and recover earlier states through historical lookback. 
These aspects capture the core challenges of memory management in realistic long-horizon interactions and enable systematic analysis of which part of memory capabilities succeed or fail, beyond measuring end-to-end task success alone.

\subsection{Four Domains in \benchmark}
\label{app_benchmark: domains}
\textsc{(1) \babi} (State Tracking).
We build on bAbI~\citep{weston2015towards}, adopting its fact-based, state-tracking format with simple, compositional sentences, where each input unit corresponds to an individual fact describing an entity state.
The information is structured as discrete, symbolic facts, and updates occur through sequential, localized modifications that often explicitly overwrite previous states.
This domain, therefore, requires systems to integrate sequential updates, track precise state transitions, and perform temporal reasoning to accurately recover both current and historical states. 

\textsc{(2) HorizonBench} (Dialogue-based Multi-turn Interactions).
Based on HorizonBench~\citep{li2026horizonbench}, a long-horizon personalization benchmark with simulated users and 6-month conversation histories,
we construct multi-turn dialogue contexts, where each input unit is a dialogue session composed of multiple conversational turns. 
Information is distributed across natural language utterances and is often implicitly expressed through user interactions. 
Updates are incremental, noisy, and indirect, requiring models to interpret evolving user intent and preferences over time. 
This domain evaluates whether memory systems can maintain and update such implicit changes over time through the conversation and answer questions about the resulting user state.

\textsc{(3) Wiki Revisions} (Factual Knowledge QA).
We construct contexts from Wikipedia revision histories. Each input instance consists of a single article paired with its full chronological sequence of revisions, where each revision is a complete document snapshot augmented with metadata, e.g., timestamp, editor identity, and edit comment. This setting differs from single-snapshot or synthetic memory benchmarks in that it exhibits substantial temporal heterogeneity. Facts may be added, refined, contradicted, or removed over time; sections may be reordered; and a given attribute typically assumes a sequence of values rather than a single fixed value.
Consequently, answering a query requires reconstructing a prior state of the article, identifying which editor introduced a claim, counting the number of value changes, or distinguishing outdated information from currently valid content. A memory system, therefore, must preserve revision-level provenance, track the evolution of attributes across revisions, and differentiate superseded information from information that remains valid.

\textsc{(4) Git Commits} (Code and Files Evolution).
We construct contexts from GitHub commit histories in an analogous manner. Each input instance consists of a single repository paired with its full chronological sequence of commits, where each commit is a complete snapshot of the codebase augmented with metadata, e.g., author, timestamp, commit message, and the set of modified files. The requirements introduced in the Wikipedia setting, i.e., preserving provenance, tracking the evolution of attributes, and distinguishing outdated from currently valid information, transfer directly to this domain.
A key distinction is that each snapshot comprises structured, executable code rather than prose, and therefore specifies not only a textual state but also concrete program behavior. This gives rise to phenomena that are largely absent in natural-language histories. First, edits are often cross-file and tightly coupled, e.g., a single commit may rename a function and update all corresponding call sites. Second, the same identifier, e.g., a function name, API signature, or configuration key, may assume a sequence of distinct semantics over time. As a result, a memory system operating in this setting must additionally recover the implicit differences between successive snapshots and reason about how program behavior evolves across commits.

\subsection{Question Examples for Each Domain}
\label{app_benchmark: q_example}
In Table~\ref{tab:example-questions-by-category}, we provide the question examples for each domain and question category.

\begin{table*}[t]
\caption{Example questions for each domain and category.}
\label{tab:example-questions-by-category}
\centering
\small
\begin{tabular}{@{}l l >{\arraybackslash}m{0.62\linewidth}@{}}
\toprule
\textbf{Domain} & \textbf{Category} & \textbf{Example Question} \\
\midrule
\multirow{5}{*}[2.5\baselineskip]{Wiki}
  & \simple    & In what year did the article first mention that Toni Basil provided choreography for the tour? \\
  & \history   & How does the version of the article 42 edits before the latest version format the team name for the No. 1 car in the race classification table? \\
  & \counting & How many distinct numerical peak chart positions has the article ever listed for the single 'Uh Huh', up to the current version? \\
  & \order& Among all the phrases the article has used to describe what Conan's decapitation of Thulsa Doom revealed him to be, what was the longest span (in days) that any single phrase was listed? \\
  & \multihop & At the revision immediately before the article updated the tree's maximum height from 20 metres to 50 m, what taxonomic division was the tree classified under? \\
\midrule
\multirow{5}{*}[2.5\baselineskip]{GitHub}
  & \simple    & Which contributor first introduced the \texttt{tests/admin/test\_api\_revoking\_admin\_role.py} file into the project? \\
  & \history& What value does the version of the project 70 commits before the latest version set for \texttt{inference} in the \texttt{[yolo2]} section of \texttt{config.ini}? \\
  & \counting& How many distinct version strings has the project ever set in \texttt{setup.py}, up to the current version? \\
  & \order& For how many days was \texttt{MAX\_DRAFT\_SENTENCES} set to \texttt{5} in \texttt{manager.py} before it was increased? \\
  & \multihop & Just before the default \texttt{arm\_velocity\_limit} parameter was modified for the second time in the arm controller, what network interface was set as the default for the loco client? \\
\midrule
\multirow{5}{*}[17\baselineskip]{HorizonBench}
  & \simple    & What is the user's current value for `preferred source types' in their ``Empirical Evidence Integration Style'' preference? \\
  & \history & What was the user's value for `preferred response format' in their ``Collaborative Alternate-History Storytelling Interaction Style'' preference 4 preference-change events ago? \\
  & \counting  & How many times has the user changed their value for `encouragement tone preference' in their ``Self-Esteem Rebuilding Communication Style'' preference? \\
  & \order  & List in chronological order all distinct values the user has held for `language register' in their ``Self-Esteem Rebuilding Interaction Style'' preference, from earliest to most recent. Output as a comma-separated list. \\
  & \multihop & For the user's `openness to nonwestern' in their ``Philosophical Tradition Affinity'' preference, which value have they held longer: \texttt{very\_high} or \texttt{high}? \\
\midrule
\multirow{5}{*}[2\baselineskip]{bAbI}
  & \simple    & Who last dropped football? \\
  & \history   & Where was Daniel for the sixteenth most recent time? \\
  & \counting  & How many total times has milk been picked up? \\
  & \order  & List in chronological order all distinct people who have dropped the milk, from earliest to most recent. Output as a comma-separated list. \\
  & \multihop & Who most recently traveled directly from office to kitchen? \\
\bottomrule
\end{tabular}
\end{table*}

\subsection{Question Generation}
\label{app_benchmark: q_generation}
For both \babi and \horizonbench, we generate questions using the provided metadata or parsed facts using predefined question templates. 
In the \babi setting, a subset of the \simple, \counting, and \multihop questions is adopted from OAKS-BABI~\citep{kim2026largelanguagemodelsup}. 
We additionally construct new questions to align with our task definitions. 
For the remaining questions, similar to OAKS-BABI construction, we parse each fact into a structured triplet of (subject, object, verb) and instantiate the question template using the parsed information.
For \horizonbench, we use the provided metadata\footnote{\url{https://huggingface.co/datasets/stellalisy/HorizonBench}}, which contains information such as user preferences. Similar to the \babi setup, we design a template for each question type and populate it with the corresponding metadata fields. Since the exact answer words may not explicitly appear in the context and are only available in the metadata, we provide candidate options together for those questions. 

For Wiki-Revisions and Git Commits, we use the official APIs to collect revision histories of articles and repositories, respectively. They are obtained from the MediaWiki\footnote{\url{https://en.wikipedia.org/w/api.php}} and GitHub\footnote{\url{https://api.github.com}} APIs. 
For Wikipedia, we restrict candidate articles to those in the Featured Articles and Good Articles categories, i.e., Wikipedia’s community-curated and peer-reviewed quality tiers. In addition, we require each article’s current size and prose density to fall within a predefined range, excluding stubs, list pages, and pages dominated by templates or infoboxes. For GitHub, we limit our selection to non-forked and non-archived Python repositories with at least 100 stars to ensure quality. 
In both domains, we keep samples that contain a sufficient number of substantive revisions up to 100. This ensures that each sample provides adequate temporal depth for probing memory evolution. We also remove non-substantive edits, e.g., bot-generated changes, markup-only updates, or empty revisions, so that each retained revision reflects a meaningful modification.
We then generate questions using Gemini-3.1-Pro~\citep{gemini31pro} with descriptions and examples of each question type and the complete revision history under a structured output schema to generate questions. 
Specifically, for Wiki Revision, the article's earliest version, followed by every subsequent revision with revision metadata (\texttt{revision\_ids}, \texttt{timestamp}, \texttt{editor}, \texttt{edit\_comment}) are provided to Gemini-3.1-Pro. Each generated question is paired with the \texttt{revision\_ids} that serve as supporting evidence.
Similarly, for Git Commit, the repository's oldest captured commit, followed by every subsequent commit as that commit's combined multi-file unified diff against its parent, each augmented with commit metadata (\texttt{timestamp}, \texttt{username}, \texttt{commit\_message}) are given to Gemini-3.1-Pro. 

The templates and prompts used in this work are included the official GitHub (\url{https://github.com/amy-hyunji/MINTEval}) due to their length.

\subsection{Human Validation on the Generated Data}
We further conduct a human validation on 405 stratified samples drawn from both the Wiki Revisions and Git Commits subsets, covering five question types. 
We find that 95.6\% of the samples are valid, meaning that both the question and answer are correctly annotated. 
Only a small proportion of cases are invalid, including 1.0\% when both question and answer are invalid, 1.7\% where only the question is invalid, and 1.7\% where the answer is only invalid. Breaking down the results by dataset, Git Commits exhibits a 98.0\% validity rate, whereas Wiki Revisions shows a slightly lower but still strong validity rate of 93.2\%. Across question categories, \simple questions show the highest validity of 97.6\%, \history questions show 93.9\%, \counting show 93.8\%, \order show 97.5\%, and \multihop show 95.0\%.  
Counting and ordering tasks are fully valid, while \simple, \history, and \multihop questions show moderately lower validity (86.7\%, 80.0\%, and 81.8\%, respectively), suggesting that more complex queries are more prone to annotation issues. Overall, these results indicate that the dataset is generally reliable, with errors concentrated in more complex question types.

\subsection{Dataset Statistics}
\label{app_benchmark: stats}
\begin{table*}[t!]
\centering
\caption{
Dataset statistics across four domains.
\textit{Depth} denotes the number of turns, revisions, or commits in each example.
\textit{k} indicates values reported in thousands.
}
\begin{tabular}{cl|cccc}
\toprule
 && \babi & HorizonBench & Wiki Revision & Git Commit \\
\midrule
&Domain   & State Tracking & Dialogue & Wikipedia & Code \\
&\# Sessions & 99 & 100 & 196 & 200 \\
\midrule
\multirow{4}{*}{\shortstack{\textit{Context}\\\textit{Statistics}}}
&Depth (avg) & 42  & 142 & 99  & 61 \\
&Depth (max) & 148 & 183 & 100 & 100 \\
&Tokens (avg) & 0.3k  & 274k  & 195k  & 86k \\
&Tokens (max) & 0.9k  & 496k  & 1768k & 600k \\

\midrule
\multirow{6}{*}{\shortstack{\textit{Question}\\\textit{Distributions}}}
&Simple   & 720  & 998  & 319 & 283 \\
&History  & 1936 & 2909 & 524 & 602 \\
&Ordering & 1000 & 1000 & 247 & 305 \\
&Counting & 1000 & 998  & 63  & 128 \\
&Multihop & 1000 & 1000 & 339 & 260 \\
\cmidrule{2-6}
&\# Total  & 5656 & 6905 & 1492 & 1578 \\
\bottomrule
\end{tabular}
\label{table: Q_stats}
\end{table*}

We provide more detailed statistics in Table~\ref{table: Q_stats}. Across domains, the contexts vary substantially in both depth and total token length, ranging from short synthetic trajectories to highly long-form histories exceeding one million tokens. The benchmark also contains a balanced distribution of question types, including simple recall, historical lookup, ordering, counting, and multihop reasoning, enabling systematic evaluation of memory retrieval, temporal reasoning, and aggregation capabilities under interference-heavy contexts.

\section{More Experimental Details}
\label{app_exp: setup}
For all experiments, we set the decoding temperature to 0. 
Models are instructed to present the final answer wrapped in \texttt{\textbackslash boxed\{\}}. 
We conduct experiments on a server either with 4x 80GB A100 or 4x 48GB A6000.

\section{Additional Analysis}
\subsection{Impact of Answering Agent Choice}
\label{app: memagent_answering_agent}

\begin{figure*}[t!]
    \centering
    \includegraphics[width=\linewidth]{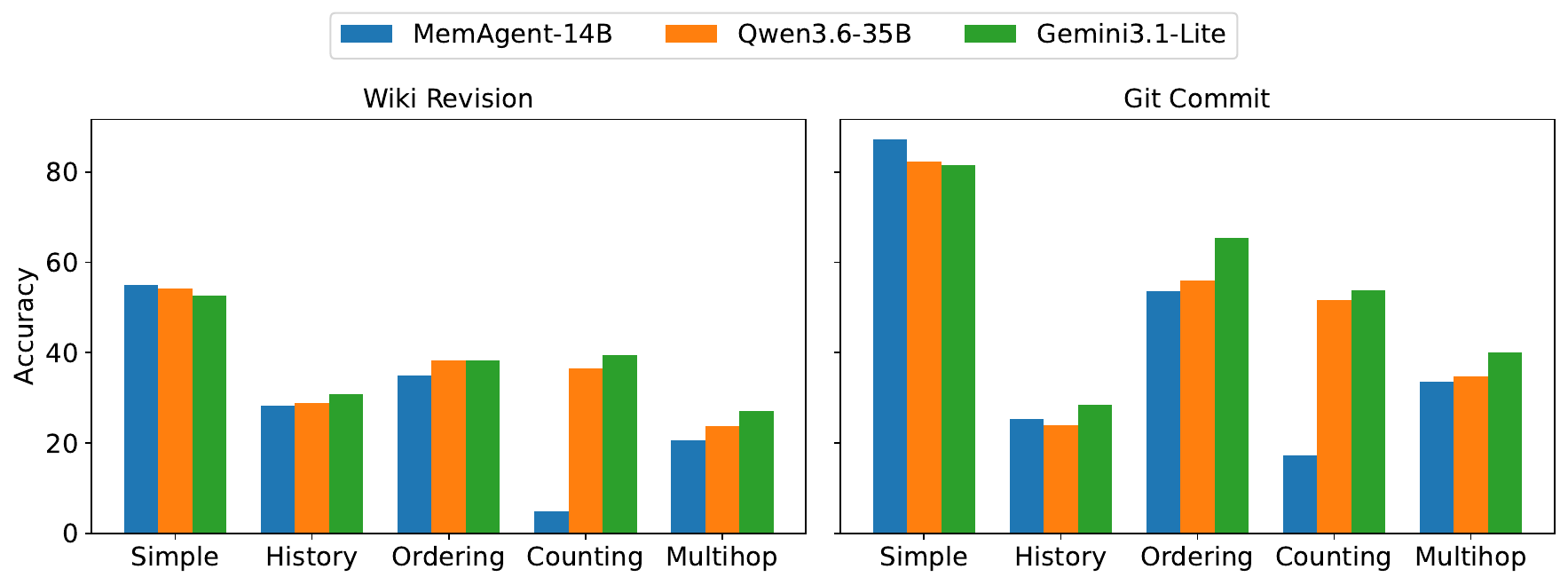}
    \caption{
    MemAgent performance on Wiki Revisions and Git Commits across different answering agents. Specialized answering agents such as MemAgent-14B remain competitive on single-target recall, but drop on multi-target aggregation questions (especially \counting), which require stronger aggregation and reasoning capabilities.
    }
    \label{app_fig: memagent_answering_agent}
\end{figure*}

Figure~\ref{app_fig: memagent_answering_agent} shows the performance of MemAgent~\citep{yu2025memagentreshapinglongcontextllm} paired with different answering agents, including the originally trained MemAgent-14B, \qwen, and \gemini. 
We observe that when experimenting with MemAgent-14B, a smaller but specialized answering agent, the overall performance remains competitive on \textit{single-target recall}, but drops on \textit{multi-target aggregation} questions, especially on \counting questions, which require stronger aggregation and reasoning capabilities.

\subsection{Using Frontier Models with the Full Context Remains Competitive}

\begin{figure*}
    \centering
    \includegraphics[width=\linewidth]{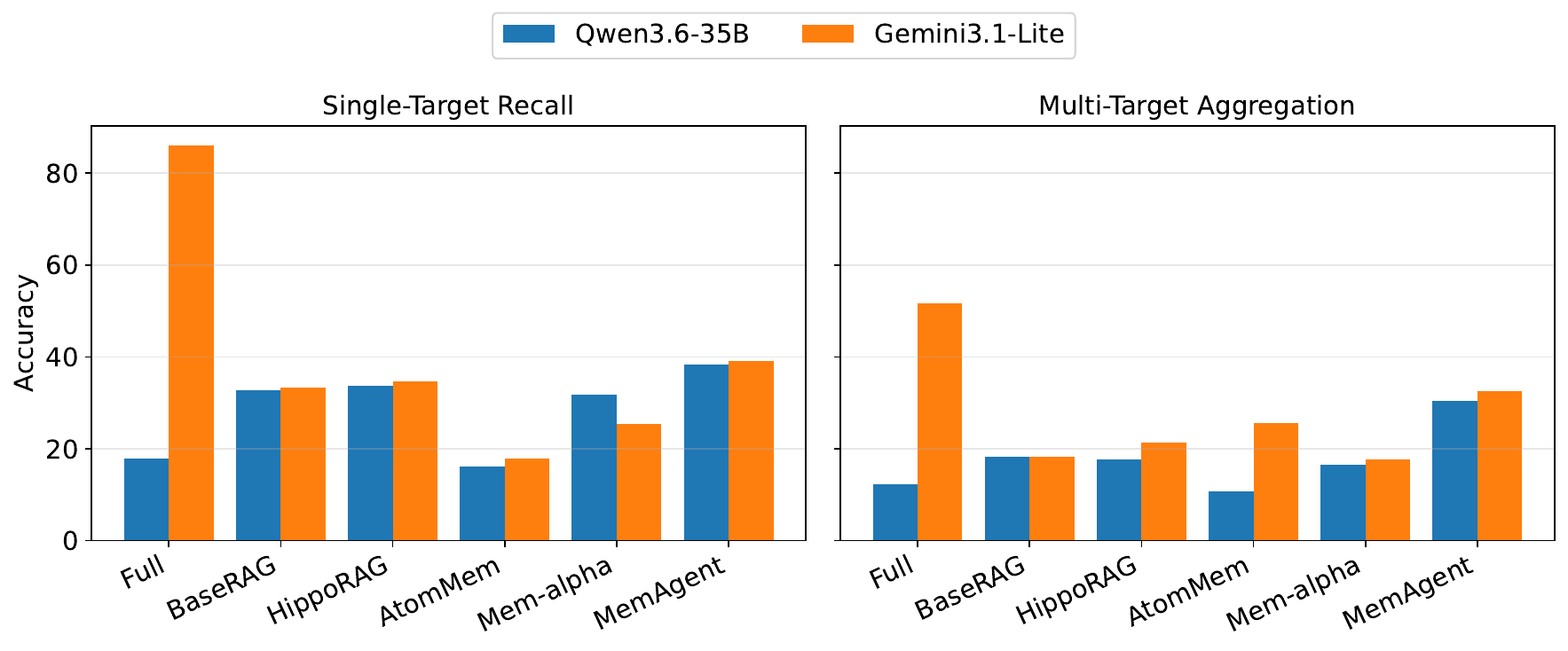}
    \caption{
    Comparison of performance across different answering agents (\qwen and \gemini). 
    The performance gap is largest under the Full Context setting. 
    Overall, the gap is larger on Multi-Target Aggregation tasks than on Single-Target Recall task. 
     }
    \label{app_fig: gemini_qwen_comparison}
\end{figure*}
In Figure~\ref{app_fig: gemini_qwen_comparison}, we compare the performance of different methods when using \qwen and \gemini as answering agent. Using \gemini with the Full Context shows the highest performance on both \textit{single-target recall} and \textit{multi-target aggregation} tasks. The improvement is particularly pronounced for \textit{single-target recall}, where \gemini with Full Context achieves over 80\% accuracy, far surpassing other retrieval-based and memory-augmented systems. 
These results suggest that frontier models like \gemini not only support longer context length, but can also effectively reason over long and interference-heavy contexts.
However, once retrieval or memory modules are introduced, the performance gap between \qwen and \gemini becomes relatively small. This indicates that, in memory-augmented settings, the quality of the context, i.e., retrieved content or memory, is important.

\subsection{Effect of Adding Temporal Cues to \history Questions}
\label{app_exp: adding_date_and_time}

\begin{figure}
    \centering
    \includegraphics[width=0.5\linewidth]{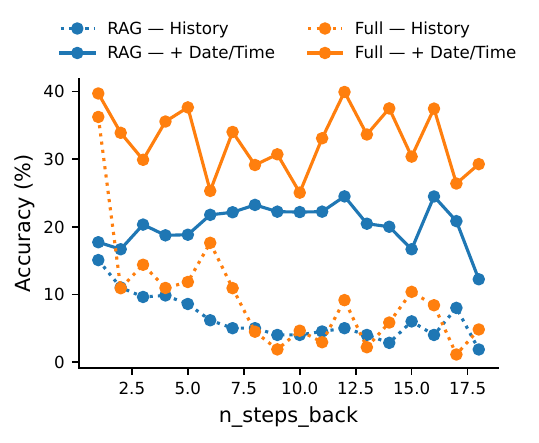}
    \caption{Performance on \history questions in \babi as a function of lookback distance (x-axis), comparing RAG and Full Context methods with and without temporal cues (\textit{History} vs. \textit{+Date/Time}). Adding timestamps as explicit markers helps recover the gap caused by interference.}
    \label{fig:add_datetime}
\end{figure}

To investigate whether the performance degradation with increasing lookback distance in Figure~\ref{fig: n_step_back} is caused by interference among similar facts, we conduct an additional experiment in which we add explicit cues (date and time information) to both the facts and the questions. 
These cues help distinguish otherwise similar facts and make them more discrete. We perform this experiment on \babi, where such cues can be easily incorporated into the data generation process. 
Figure~\ref{fig:add_datetime} compares performance with and without datetime information under the same inputs and questions. 
We observe that adding these cues substantially mitigates the performance degradation as the lookback distance increases for both Full Context and RAG systems. In contrast, without the cues, performance drops sharply as the distance increases.

\subsection{Biased Toward Insertion in Memory Systems}
\label{app: tool_usage_rate}
\begin{figure*}
    \centering
    \includegraphics[width=\linewidth]{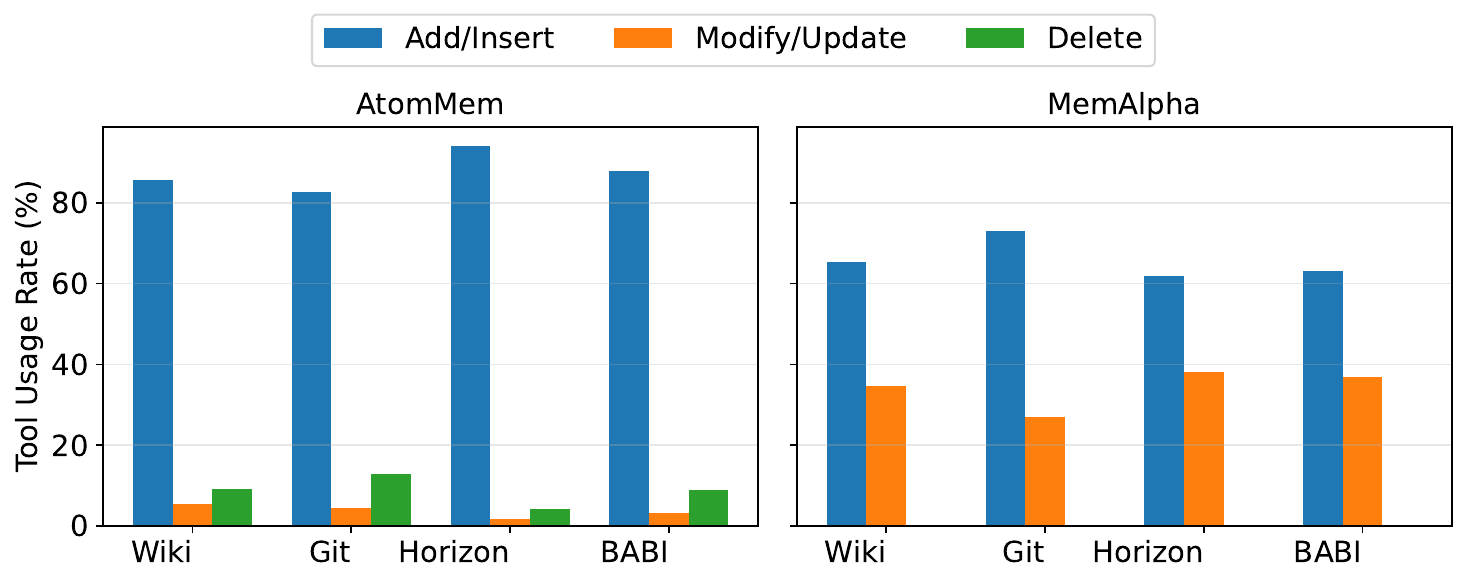}
    \caption{
    Rate of tool usage for AtomMem and Mem-$\alpha$. Mem-$\alpha$ consistently underutilizes the delete operation across all datasets, which may partially explain why memory systems struggle in long-horizon settings with heavy interference: outdated or conflicting information accumulates over time, leading to progressively greater conflict within memory.
    }
    \label{app_fig: tool_usage_rate}
\end{figure*}

In Figure~\ref{app_fig: tool_usage_rate}, we analyze the distribution of three operations: (1) inserting new information, (2) modifying or updating existing entries, and (3) deleting outdated information.
Comparing the two systems, Mem-$\alpha$ demonstrates a substantially higher rate of modification operations (34.1\%) than AtomMem (3.7\%), indicating a better ability to update existing memory instead of duplicating it, suggesting why Mem-$\alpha$ shows stronger overall performance. However, Mem-$\alpha$ tends to underutilize the delete operation across all datasets, which could partially explain why memory systems fail under long-horizon settings with heavy interference, as outdated or conflicting information accumulates over time and increases conflicting information.

\subsection{Effect of Retrieval Choices on RAG Performance}
\label{app: RAG}

\begin{figure}[t!]

    \begin{minipage}[t]{0.48\linewidth}
    \centering
    \includegraphics[width=\linewidth]{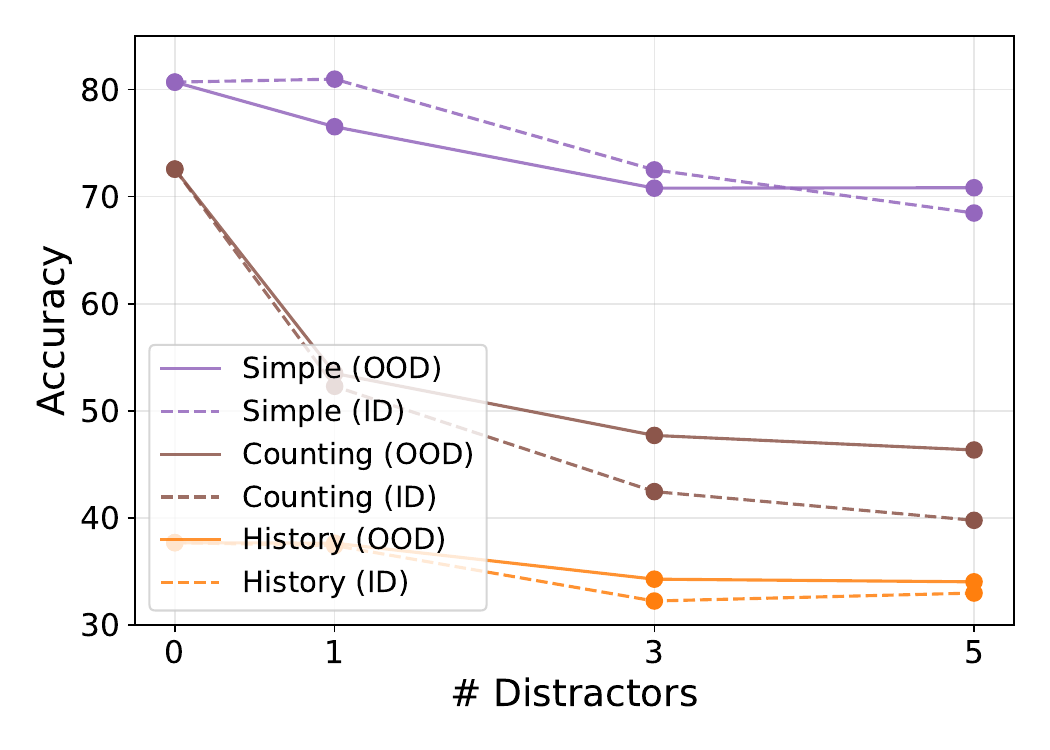}
    \caption{Performance under varying distractor types and numbers of distractors. \textit{ID} distractors more strongly affect questions such as \counting and \history compared to simpler queries like \simple, suggesting that tasks requiring aggregation or tracking over multiple facts are more susceptible to interference.}
    \label{app_fig: distracting}
    \end{minipage}
        \hfill
    \centering
    \begin{minipage}[t]{0.48\linewidth}
    \centering
    \includegraphics[width=\linewidth]{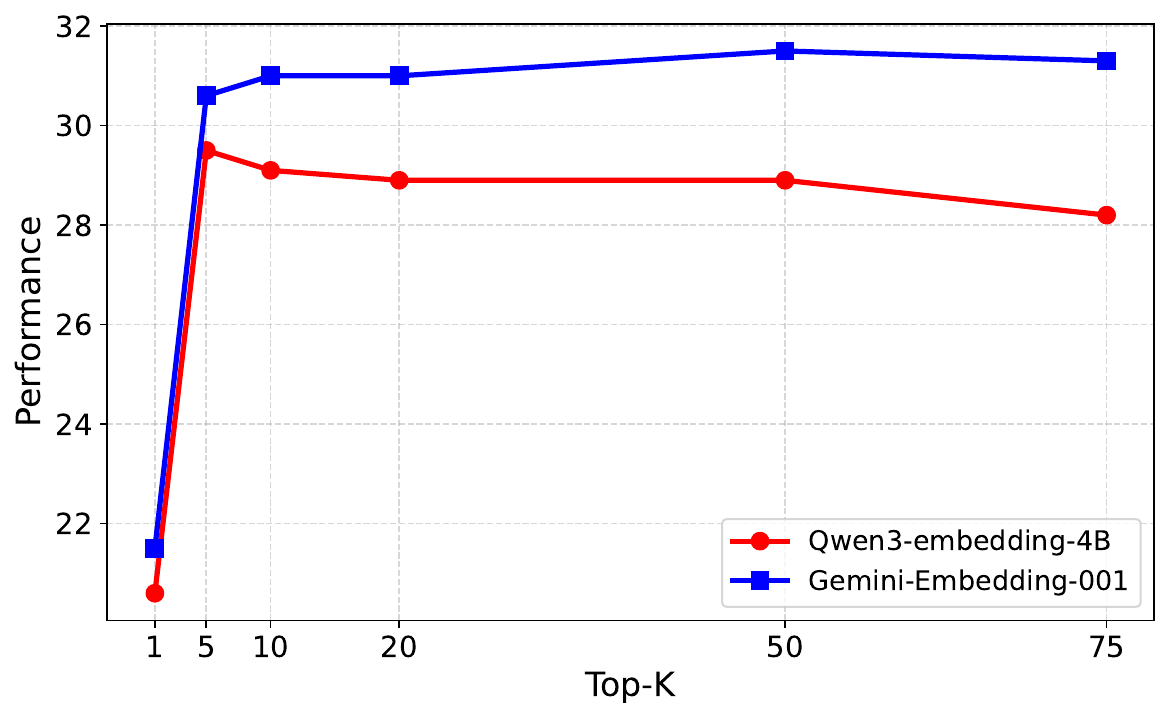}
    \caption{
    Average performance across all datasets for different retrieval models (Qwen3-Embedding-4B and Gemini-Embedding-001) as the number of retrieved documents varies, using \qwen as the answering agent.
    } 
    \label{fig: retrieval_topk}
    \end{minipage}

\end{figure}

We analyzed how retrieval design choices—specifically the embedding model and the number of retrieved documents ($K$)—affect downstream question-answering performance in a RAG setting. Experiments are conducted over average of all four datasets using RAG, while keeping the answering model fixed as \qwen. We compare two embedding models: Qwen3-Embedding-4B~\citep{qwen3embedding} and Gemini-Embedding-001~\citep{geminiemb}.

As shown in Figure~\ref{fig: retrieval_topk}, average performance increases sharply from $K=1$ and $K=5$, after which gains largely plateau. Qwen3-Embedding-4B achieves its best performance at $k=5$, while Gemini-Embedding-001 peaks around $K=50$, though performance remains relatively similar for larger $K$ values overall.
When comparing retrieval models, Gemini-Embedding-001 consistently outperforms Qwen3-Embedding-4B over all $K$ values, with the performance gap widening slightly as $K$ increases. This indicates that stronger embeddings are more effective at ranking relevant documents higher when the retrieval pool is larger. 

\begin{figure*}[t!]
    \centering
    \includegraphics[width=\linewidth]{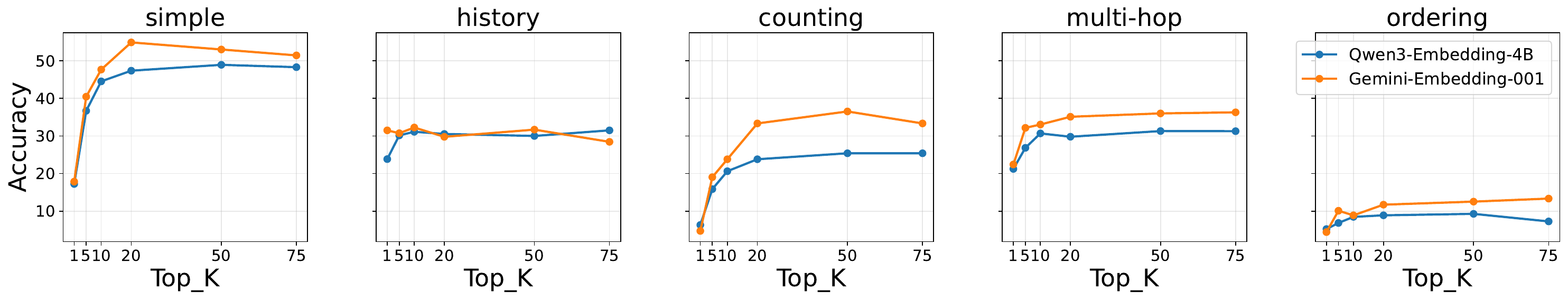}
    \caption{
    RAG performance across question types with varying numbers of retrieval documents and embedding models (Qwen3-Embedding-4B and Gemini-Embedding-001).
    }
    \label{app_fig: qtype_RAG}
\end{figure*}

A finer-grained analysis by question type~(Figure~\ref{app_fig: qtype_RAG}) on Wiki Revision dataset reveals that most of the performance gap between embedding models arises from more complex \textit{multi-target aggregation} questions, especially \counting and \order questions. We hypothesize that this is because these question types typically require aggregating or comparing information across multiple pieces of evidence. Increasing $K$ leads to a higher probability that all necessary evidence is retrieved, which disproportionately benefits these reasoning-heavy categories. In contrast, simpler \textit{single-target recall} type questions (i.e., \simple or \history) show smaller sensitivity to both embedding choice and retrieval depth, as they often depend on retrieving a single highly relevant document.

\subsection{Expanded Discussion on the State-of-the-art Memory System Failure}
\label{app_exp: simplemem}

\begin{table}[t]
\centering
\small
\caption{Results on SimpleMem using \gemini and Gemini-Embedding-001 across datasets and question types, reported in Exact Match (\%). Even the SOTA memory system, combined with frontier models, struggles on \benchmark.}
\label{tab:gemini_simple_mem}
\begin{tabular}{lccccc|c}
\toprule
\textbf{Dataset} & \textbf{Simple} & \textbf{History} & \textbf{Counting} & \textbf{Ordering} & \textbf{Multi-hop} & \textbf{Overall} \\
\midrule
\babi          & 93.2 & 48.9 & 74.8 & 73.8 & 52.3 & 67.7 \\
\horizonbench  & 6.3  & 5.7  & 10.8 & 3.1  & 23.5 & 8.8  \\
Wiki Revisions  & 7.2  & 20.4 & 31.8 & 0.0  & 11.8 & 12.7 \\
Git Commits    & 83.0 & 13.1 & 25.0 & 30.5 & 26.5 & 32.2 \\
\bottomrule
\end{tabular}
\end{table}

SimpleMem is a state-of-the-art memory architecture built around a three-stage pipeline:
(1) \textit{Semantic Structured Compression}, which distills unstructured interactions into compact multi-view memory units;
(2) \textit{Online Semantic Synthesis}, which incrementally merges related contexts into unified abstractions to reduce redundancy; and
(3) \textit{Intent-Aware Retrieval Planning}, which dynamically infers retrieval scope and constructs targeted retrieval contexts.
Using frontier models, \gemini and Gemini-Embedding-001, we successfully reproduced the reported results on LoCoMo~\citep{maharana2024evaluating}, achieving a state-of-the-art F1 score of 54.76\%.
As shown in Table~\ref{tab:gemini_simple_mem}, performance degrades dramatically on \benchmark.
The failure arises from a fundamental mismatch between the assumptions underlying conversational memory benchmarks and the characteristics of revision-centric data. In LoCoMo, each turn contains roughly 109 characters, yielding approximately 4.4k characters in a memory chunk. Compressing this context into 5--10 structured memory entries is therefore feasible with limited information loss. In contrast, our benchmark contains revisions with a median length of 4.6k characters. A memory chunk consequently expands to approximately 184k characters. Compressing such a window into the same 5--10 memory entries discards the majority of the source content.
Moreover, the compression objective itself is actively harmful in this setting. SimpleMem explicitly encourages the model to avoid duplication during memory construction. This assumption is appropriate for dialogue, where repeated statements are often redundant, but it is detrimental for revision histories. In our dataset, consecutive revisions exhibit substantial lexical overlap, while the critical information often lies in small localized edits. We observe that the performance drops much more on \horizonbench, Wiki Revisions and Git Commits, as revision provenance is not retained through the compression pipeline (i.e., has been paraphrased or rewritten). As no explicit metadata records which revision produced a given fact, retrieval operates solely over keywords and embeddings, making queries such as retrieving the contents of ``Revision 53'' more challenging.
We also experimented with \qwen and Qwen3-4B retrieval model, but observed near-zero performance across all datasets; therefore, we do not report the results.
An interesting future direction for mitigating memory interference is to distribute memory management across specialized agents~\citep{lin2026squidagentparallelizewiselycoordinate, yun2026graphofagentsgraphbasedframeworkmultiagent}.

\begin{table}[t]
\centering
\small
\caption{
Accuracy (\%) on state tracking dataset by the number of facts observed. 
We evaluate each method after every five observed facts and score only questions that can be answered from the facts seen so far. 
Accuracy declines across all methods as more facts are observed.
}
\label{tab:online_eval}
\begin{tabular}{lcccccc}
\toprule
\textbf{Facts Observed} & \textbf{$\leq$ 10} & \textbf{11-20} & \textbf{21-30} & \textbf{31-50} & \textbf{51-80} & \textbf{81+} \\
\midrule
Full & 98 & 83 & 83 & 79 & 64 & 44 \\
RAG & 70 & 59 & 54 & 43 & 36 & 26 \\
HippoRAG & 78 & 33 & 22 & 16 & 10 & 6 \\
Mem-$\alpha$& 95 & 72 & 69 & 64 & 48 & 41\\
MemAgent & 84 & 74 & 71 & 65 & 53 & 51\\
\bottomrule
\end{tabular}
\end{table}

\subsection{Online Evaluation Performance}
\label{app: online_eval}

We evaluate memory systems online in the state-tracking domain, where we could recompute gold answer as new facts arrive using it's synthetic structure.
We retain the original questions and evaluate each method after every five observed facts.
At each evaluation, we recompute the gold answers using only the facts observed so far and report accuracy over the questions that can be answered at that point.
As shown in Table~\ref{tab:online_eval}, accuracy declines across all evaluated methods as the interaction horizon grows. From the shortest range (at most 10 facts) to the longest range (at least 81 facts), accuracy falls from 98\% to 44\% for Full Context, from 95\% to 41\% for Mem-$\alpha$, and from 70\% to 26\% for RAG. 
History questions show the largest declines. Between horizons of at most 15 facts and at least 81 facts, accuracy on History questions falls from 94\% to 9\% for Full Context and from 83\% to 20\% for Mem-$\alpha$, an average drop of 74 percentage points.
This pattern is consistent with our offline finding in Section~\ref{sec4: lookback} that earlier states become harder to recover after many subsequent updates. 
Accuracy also drops more sharply for lossy memory systems on queries issued immediately after a state change, falling from 62.0\% to 30.3\% for RAG and from 80.4\% to 60.7\% for Mem-$\alpha$. Full Context shows a smaller drop, from 95.7\% to 87.0\%. These results suggest that interference from state updates contributes to errors in lossy memory systems. Overall, the agreement between the online and offline results suggests that the offline protocol captures key patterns of memory degradation at lower computational cost.

\subsection{Evaluating Partial Correctness with an LLM Judge}
\label{app: partial_correctness}

\begin{table}[t]
\centering
\small
\caption{
Average EM and LLM judge scores with partial correctness by question type. 
}
\label{tab:llm_judge}
\begin{tabular}{lccc}
\toprule
\textbf{Question Type} & \textbf{Avg. EM} & \textbf{Avg. LLM Judge}\\
\midrule
\simple & 47.5 & 49.5 \\
\counting & 27.3 & 35.6 \\
\history & 21.0 & 28.9 \\
\order & 24.0 & 34.7 \\
\multihop & 28.0 & 31.1 \\
\bottomrule
\end{tabular}
\end{table}

For additional insight beyond EM, we conduct partial correctness and reasoning process using an LLM Judge with a five-level scoring rubric. 
As shown in Table~\ref{tab:llm_judge}, simple questions show little difference as they typically require a single fact. In contrast, the largest improvement occurs for ordering questions, as predictions with minor ordering errors (e.g., swapping two adjacent events) receive partial credit, followed by history and counting, where answers are often largely correct but miss only a small number of entities or historical states.

\subsection{Training with \benchmark}
\label{app: train}
\begin{table}[t]
\centering
\small
\caption{
Performance on \benchmark and HotpotQA with full-context inference (Full), retrieval-augmented generation (RAG), and a model trained on the bAbI subset of \benchmark.
}
\label{tab:training}
\begin{tabular}{l|c|c|ccc}
\toprule
 & \multicolumn{1}{c|}{In-Domain} & \multicolumn{4}{c}{Out-of-Domain}\\
 \midrule
 & bAbI & HotpotQA & Wiki Revisions & Git Commits & HorizonBench\\
\midrule
Full & 21.8 & 31.4 & 10.2 & 10.8 & 6.5\\
RAG & 36.4 & 45.1 & \textbf{19.5} & \textbf{23.5} & 5.6 \\
Ours (Trained) & \textbf{62.9} & \textbf{69.0} & 11.3 & 13.1 & \textbf{7.8} \\
\bottomrule
\end{tabular}
\end{table}

We investigate whether \benchmark can also serve as training dataset for strong memory systems. 
To this end, we create an additional training split of 1.9K state-tracking dataset and use it to train Qwen3-1.7B~\citep{qwen3technicalreport} following Mem-$\alpha$~\citep{wang2025memalpha} training procedure.
As shown in Table~\ref{tab:training}, the trained model outperforms both 1.7B baselines (Full Context and RAG) on the in-domain state-tracking evaluation and on other widely used benchmark, HotpotQA~\citep{yang2018hotpotqadatasetdiverseexplainable}. 
These results suggest that \benchmark can serve not only as an evaluation benchmark, but also as a valuable source of training data for developing stronger memory systems that generalize to practical downstream tasks.
However, it transfers only marginally to the remaining \benchmark domains (Wiki Revision and Git Commits), despite their shared data generation pipeline and similar question types. These observations suggest that simply exposing models to \benchmark-style training data is insufficient to fully close the performance gap, indicating that \benchmark measures challenging long-horizon memory capabilities.

\section{Additional Related Work}
\label{app: related_work}
\paragraph{Memory-Augmented Agents.} Memory-augmented agents span several paradigms. RAG-based approaches such as HippoRAG~\citep{gutierrez2025ragmemorynonparametriccontinual} organize extracted knowledge into graphs for associative multihop retrieval.
Among pipeline-based systems, MemGPT~\citep{packer2024memgptllmsoperatingsystems} manages OS-inspired hierarchical memory tiers via a controller that pages information in and out of context, while SimpleMem~\citep{liu2026simplememefficientlifelongmemory} maintains selectively-pruned running summaries. 
Among training-based approaches, MemAgent~\citep{yu2025memagentreshapinglongcontextllm} and Memory-R1~\citep{yan2025memory} use RL to learn structured write/retrieve/delete policies; MEM1~\citep{zhou2025mem1learningsynergizememory} trains memory compression jointly with reasoning via RL; and AtomMem~\citep{huo2026atommemlearnabledynamic} learns to decompose memory management into atomic CRUD operations via SFT and GRPO. Drawing on cognitive science, structured memory systems assign distinct roles to episodic and semantic memory. Mem-$\alpha$~\citep{wang2025memalpha} trains an RL agent over a multi-tier hierarchy, while REMem~\citep{shu2026rememreasoningepisodicmemory} constructs a dynamic memory graph for episodic retrieval, and SYNAPSE~\citep{jiang2026synapseempoweringllmagents} unifies episodic and semantic memory via spreading activation. 
Across these lines of work, a common assumption is that the goal of memory is to surface the most current and relevant state in response to a query. This shapes not only system design but also evaluation: models are typically assessed based on whether they return the correct answer for the latest state, while largely overlooking their ability to recall or aggregate information from earlier states. \benchmark addresses this gap by evaluating how well systems can recall and aggregate information in evolving and interference-heavy contexts.

\paragraph{Memory Evaluation in Large Language Models.} 
A variety of benchmarks have been proposed to evaluate memory systems in large language models. 
Conversational benchmarks~\citep{maharana2024evaluating, wu2025longmemevalbenchmarkingchatassistants} and QA-based benchmarks~\citep{hu2026evaluatingmemoryllmagents} evaluate retrieval and temporal reasoning, but typically involve less interconnected contexts and focus on questions about the most recent information.
Recent benchmarks such as StoryBench~\citep{wan2025storybenchdynamicbenchmarkevaluating} and RealMem~\citep{bian2026realmembenchmarkingllmsrealworld} introduce more densely interconnected contexts that naturally induce interference, but the interference events remain sparse and they still focus on the most recent information.
OAKS~\citep{kim2026largelanguagemodelsup} is the closest benchmark to \benchmark, as it also features naturally occurring interference and question answering over long-form contexts. However, as shown in Table~\ref{table: comparison}, OAKS contains substantially fewer interference events (avg. 4.7) than \benchmark (avg. 86) and does not include long-range lookback questions across multiple domains. 
Overall, \benchmark provides a broader and more challenging evaluation setting for memory systems, covering interference-heavy contexts, diverse lookback distances, and aggregation-based reasoning across multiple domains.

\section{Dataset License}
\label{app_checklist: license}
The datasets used in this work are released under permissive licenses that support open research and reproducibility. Specifically, HorizonBench~\citep{li2026horizonbench} is distributed under the Apache-2.0 license, which allows both academic and commercial use with minimal restrictions. The bAbI dataset \citep{weston2015towards} is released under the Creative Commons Attribution 3.0 (CC BY 3.0) license, which permits reuse and modification provided appropriate credit is given to the original authors. These licenses ensure that all datasets used in this study are compliant with open-access and reproducible research standards.

\end{document}